\documentclass[runningheads]{llncs}
\usepackage{graphicx}
\usepackage{amsmath,amssymb} % define this before the line numbering.
\usepackage{color}

\usepackage{url}
\usepackage{bm}
\usepackage{multirow}
\usepackage[T1]{fontenc} % for Paj{\k{a}}k

\begin{document}
\pagestyle{headings}
\mainmatter

\def\ACCV20SubNumber{775}  % Insert your submission number here

%===========================================================
\title{Deep Snapshot HDR Imaging\\Using Multi-Exposure Color Filter Array} % Replace with your title
\titlerunning{Deep Snapshot HDR Imaging}
\author{Takeru Suda \and Masayuki Tanaka \and Yusuke Monno \and Masatoshi Okutomi}
% \author{Takeru Suda \and
% Masayuki Tanaka\orcidID{0000-0002-5756-1904} \and
% Yusuke Monno\orcidID{0000-0001-6733-3406} \and Masatoshi Okutomi\orcidID{0000-0001-5787-0742}}
%
\authorrunning{T. Suda et al.}
\institute{Tokyo Institute of Technology, Tokyo, Japan\\
\email{tsuda@ok.sc.e.titech.ac.jp; \{mtanaka,ymonno,mxo\}@sc.e.titech.ac.jp}}

\maketitle

%===========================================================
\begin{abstract}
In this paper, we propose a deep snapshot high dynamic range (HDR) imaging framework that can effectively reconstruct an HDR image from the RAW data captured using a multi-exposure color filter array (ME-CFA), which consists of a mosaic pattern of RGB filters with different exposure levels. To effectively learn the HDR image reconstruction network, we introduce the idea of luminance normalization that simultaneously enables effective loss computation and input data normalization by considering relative local contrasts in the ``normalized-by-luminance'' HDR domain. This idea makes it possible to equally handle the errors in both bright and dark areas regardless of absolute luminance levels, which significantly improves the visual image quality in a tone-mapped domain. Experimental results using two public HDR image datasets demonstrate that our framework outperforms other snapshot methods and produces high-quality HDR images with fewer visual artifacts.
\end{abstract}

%===========================================================
\section{Introduction}

The dynamic range of a camera is determined by the ratio between the maximum and the minimum amounts of light that can be recorded by the image sensor at one shot. Standard digital cameras have a low dynamic range~(LDR) and only capture a limited range of scene radiance. Consequently, they cannot capture a bright and a dark area outside the camera's dynamic range simultaneously. High dynamic range~(HDR) imaging is a highly demanded computational imaging technique to overcome this limitation, which recovers the HDR scene radiance map from a single or multiple LDR images captured by a standard camera.

\begin{figure}[t!]
    \centering
    \begin{minipage}{0.9\hsize}
        \centering
        \includegraphics[width=\hsize,pagebox=cropbox,clip]{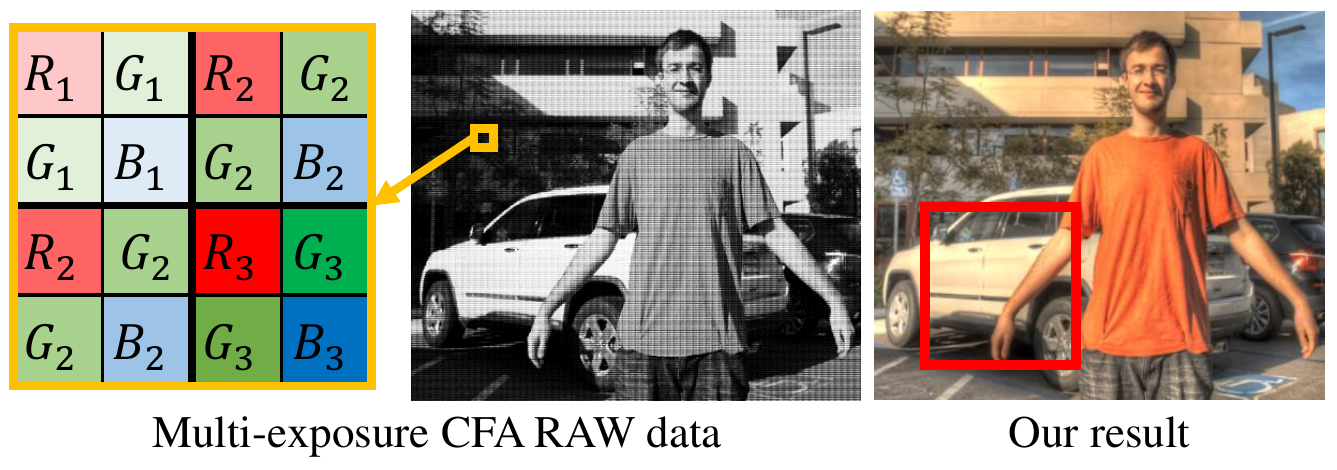}\\ 
    \end{minipage}\\
    \begin{minipage}[t]{0.22\hsize}
        \centering
        \includegraphics[width=\hsize,pagebox=cropbox,clip]{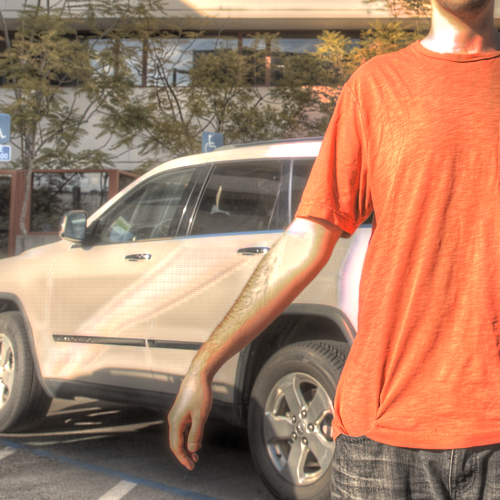}\\ 
        \footnotesize{Wu~\cite{wu2018deep}}
    \end{minipage}
    \begin{minipage}[t]{0.22\hsize}
        \centering
        \includegraphics[width=\hsize,pagebox=cropbox,clip]{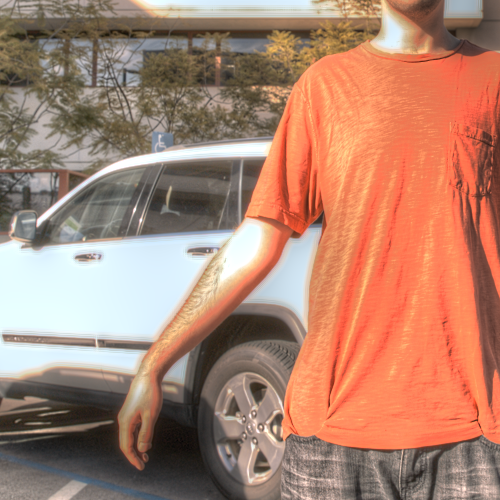}\\
        \footnotesize{ExpandNet~\cite{marnerides2018expandnet}}
    \end{minipage}
    \begin{minipage}[t]{0.22\hsize}
        \centering
        \includegraphics[width=\hsize,pagebox=cropbox,clip]{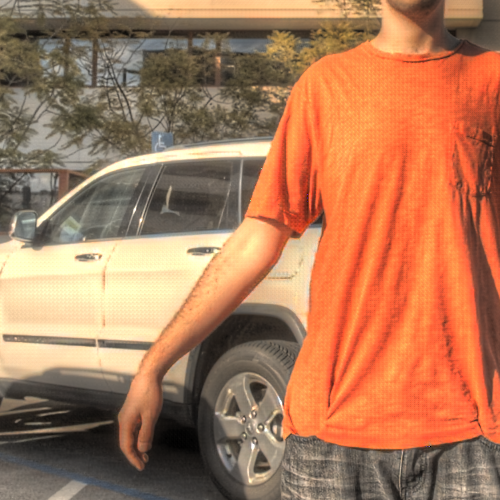}\\ 
        \footnotesize{Ours}
    \end{minipage}
    \begin{minipage}[t]{0.22\hsize}
        \centering
        \includegraphics[width=\hsize,pagebox=cropbox,clip]{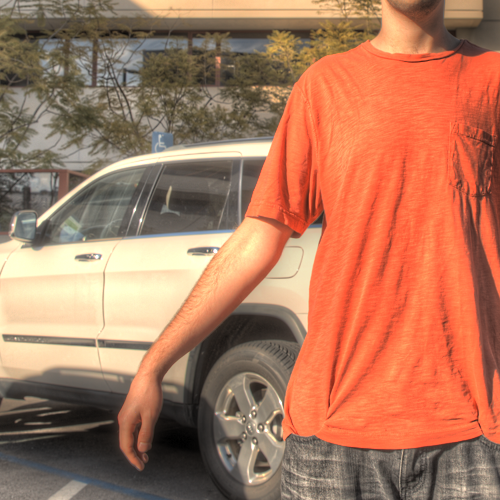}\\ 
        \footnotesize{Ground truth}
    \end{minipage}
    \caption{Top: Examples of a multi-exposure CFA RAW data and an HDR image result by our framework. Bottom: Comparisons with state-of-the-art HDR imaging methods. \label{fig:intro}}
\end{figure}

HDR imaging is typically performed by estimating a mapping from the sensor's LDR outputs to the scene radiance using multiple LDR images which are sequentially captured with different exposure levels~\cite{paul1997debevec}. Although this approach works for static situations, it is not suitable for dynamic scenes and video acquisition since the multiple images are taken at different times. Recent learning-based methods~\cite{kalantari2017deep,wu2018deep,yan2019attention} have successfully reduced ghost artifacts derived from target motions between input LDR images. However, those methods are limited to small motions and the artifacts remain apparent for the areas with large motions, as shown in Wu's method~\cite{wu2018deep} of Fig.~\ref{fig:intro}.

% /////////////////////////////////////////////////////////////////////////

Some studies have used only a single LDR image to realize one-shot HDR imaging~\cite{marnerides2018expandnet,eilertsen2017hdr,lee2018deepc}. They essentially inpaint or hallucinate missing over- and under-exposed areas by exploiting an external database of LDR-HDR image pairs. Although this approach is free from ghost artifacts, it generates inpainting or hallucination artifacts for largely missing areas, as shown in ExpandNet~\cite{marnerides2018expandnet} of Fig.~\ref{fig:intro}, due to the lack of information in the missing areas.

As another one-shot approach, the methods based on a snapshot HDR sensor have also been investigated~\cite{cho2014single,choi2017reconstructing,narasimhan2005enhancing,nayar2000high}. One way to realize a snapshot HDR sensor is to use a single image sensor with spatially varying exposure levels~\cite{narasimhan2005enhancing,nayar2000high}. This can be achieved by using what we call a multi-exposure color filter array~(ME-CFA), which consists of a mosaic pattern of RGB filters combined with neutral density filters with different attenuation levels. (see Fig.~\ref{fig:intro} for an example with three levels). The snapshot HDR sensor has the advantage of capturing multi-exposure information at one shot. Thus, it has great potential for HDR imaging of dynamic scenes and also HDR video acquisition without ghost and inpainting artifacts. However, HDR image reconstruction from the snapshot measurement includes two challenging problems: demosaicking (i.e., interpolation of missing RGB values) and HDR reconstruction (i.e., scene radiance estimation from LDR measurements). As we will show later, a simple combination of existing demosaicking/interpolation and HDR reconstruction methods cannot produce satisfactory results for this joint problem. 

In this paper, we propose a novel deep snapshot HDR imaging framework that can effectively reconstruct an HDR image from the RAW data captured using an ME-CFA (see Fig.~\ref{fig:intro}). In our framework, we introduce the key idea of luminance normalization and propose luminance-normalized network~(LN-Net) that enables effective learning of HDR images in the luminance-normalized domain~(LN domain).

In the training phase, we first train the network to generate tentatively interpolated LDR images, from which tentative HDR luminance is estimated. Then, we normalize the input ME-CFA RAW data and the corresponding ground-truth HDR image by the tentative HDR luminance. Finally, we train our main network~(LN-Net) to reconstruct HDR images based on the pair of the RAW and the HDR image data in the LN domain. In the application phase, the HDR image is reconstructed through the learned two networks. 

Our LN-Net mainly has two benefits. The first one is effective loss computation in the LN domain. The standard mean squared error~(MSE) loss in the linear HDR domain has a problem of neglecting the errors in dark areas because they are quite small compared with the errors in bright areas. However, those errors in dark areas significantly affect the visual quality in a tone-mapped domain~\cite{eilertsen2015real}, which is commonly used to display HDR images. Based on this, some studies have computed the loss in a transformed domain, such as a log domain~\cite{eilertsen2017hdr} and a global tone-mapped domain~\cite{kalantari2017deep}. However, such signal-independent transformations do not reflect an actual signal component of each image. In contrast, by computing the loss in the LN domain, we can equally handle the errors in bright and dark areas by considering the actual luminance of each image. 

The second benefit of LN-Net is effective input data normalization. In deep learning, the normalization of input data is important to extract effective features. Since a diverse range of scene radiance information is simultaneously encoded in the ME-CFA RAW data, we need to consider relative local contrasts, rather than absolute differences. Otherwise, features such as edges and textures in dark areas are prone to be ignored. In our LN-Net, by normalizing the input RAW data by the tentative luminance, we can naturally consider the relative local contrasts in both bright and dark areas regardless of absolute luminance.  

Through the experiments using two public HDR image datasets, we validate the effectiveness of our framework by comparing it with other snapshot methods and current state-of-the-art HDR imaging methods using multiple LDR images. \noindent {\bf Main contributions} of this paper are summarized as below.
\begin{itemize}
  \item We propose a novel deep learning framework that effectively solves the joint demosaicking and HDR reconstruction problem for snapshot HDR imaging.
  \item We propose the idea of luminance normalization that enables effective loss computation and input data normalization by considering relative local contrasts of each image.
  \item We demonstrate that our framework can outperform other snapshot methods and reconstruct high-quality HDR images with fewer visible artifacts.
\end{itemize}

%===========================================================
\section{Related Work}

\subsubsection{Multiple-LDR-images-based methods} have been studied for years. Their approaches include inverse radiometric function estimation~\cite{paul1997debevec}, exposure fusion \cite{ma2017multi,ma2017robust,mertens2009exposure}, patch-based~\cite{hu2013hdr,sen2012robust}, rank minimization-based~\cite{lee2014ghost,oh2014robust}, image alignment-based~\cite{hasinoff2016burst,kalantari2019deep}, and learning-based~\cite{kalantari2017deep,wu2018deep,yan2019attention,prabhakar2019fast,ram2017deepfuse,yan2019multi} methods. Although their performance has continuously been improved~(see~\cite{sen2012robust,tursun2015state} for reviews), it is essentially difficult for multi-LDR-images-based methods to handle dynamic scenes with target or camera motions, resulting in ghost artifacts. Some methods have exploited a multi-camera/sensor system~\cite{ogino2016super,tocci2011versatile} for one-shot acquisition of multi-LDR images. However, they require image or sensor alignment, which is another challenging task.

\subsubsection{Single-LDR-image-based methods,} also called inverse tone-mapping, have been actively studied in recent years. They train a mapping from a single-LDR image to an HDR image directly~\cite{marnerides2018expandnet,eilertsen2017hdr,yang2018image,moriwaki2018hybrid,kim2018itm} or train a mapping to multiple LDR images intermediately, from which the HDR image is derived~\cite{lee2018deepc,endo2017deep,lee2018deepr}. Although the single-LDR-image-based approach realizes one-shot HDR image acquisition, it is essentially difficult to reconstruct high-quality HDR images because there are no measurements obtained from different exposure levels.

\subsubsection{Snapshot methods} are based on a snapshot HDR imaging system with spatially varying exposure levels~\cite{narasimhan2005enhancing,nayar2000high}. Several hardware architectures or concepts have been proposed to realize a snapshot system, such as a coded exposure time~\cite{cho2014single,gu2010coded,uda2016variable}, a coded exposure mask~\cite{alghamdi2019reconfigurable,nagahara2018space,serrano2016convolutional}, a dual-ISO sensor~\cite{choi2017reconstructing,go2018image,hajisharif2015adaptive,heide2014flexisp}, and what we call an ME-CFA, which consists of the mosaic of RGB filters combined with neutral density filters with different attenuation levels~\cite{narasimhan2005enhancing,nayar2000high,aguerrebere2017bayesian,aguerrebere2014single,an2017single,rouf2018high,cheng2009high}. The snapshot systems have great potential for HDR imaging in dynamic situations since it enables one-shot acquisition of multi-exposure information. However, HDR image reconstruction from the snapshot measurements is very challenging due to the sparse nature of each color-exposure component.

Some existing snapshot methods based on an ME-CFA first convert the snapshot LDR measurements to the sensor irradiance domain. By doing this, the problem reduces to the demosaicking problem in the sensor irradiance domain, for which several probability-based~\cite{aguerrebere2017bayesian,rouf2018high} or learning-based~\cite{an2017single} approaches have been proposed. However, this combined approach could not necessarily produce satisfactory results because the errors in the first step are propagated by the demosaicking step. Although some joint approach has also been proposed~\cite{narasimhan2005enhancing,nayar2000high,aguerrebere2014single,cheng2009high}, it is limited to a specific ME-CFA pattern~\cite{narasimhan2005enhancing,nayar2000high,cheng2009high} or it only performs a limited over- and under-exposed pixels correction~\cite{aguerrebere2014single}.

In this paper, different from existing methods, we propose a general and high-performance framework exploiting deep learning to jointly solve the demosaicking and the HDR reconstruction problems for snapshot HDR imaging using an ME-CFA.

% /////////////////////////////////////////////////////////////////////////
\begin{figure*}[t!]
\centering
\includegraphics[width=12cm,pagebox=cropbox,clip]{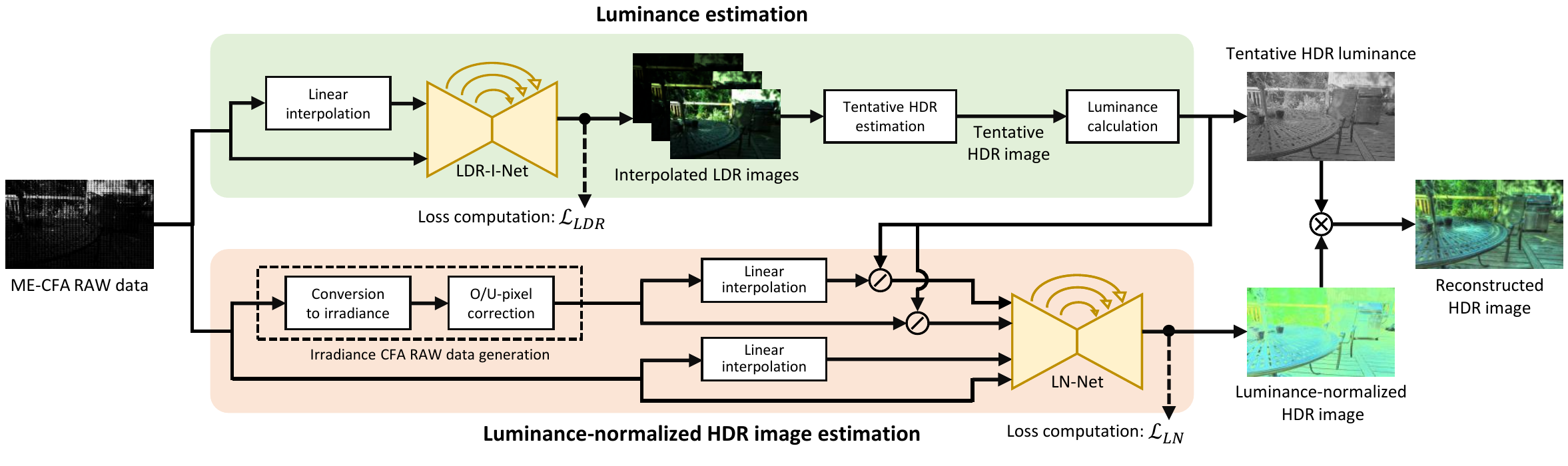}
\caption{Overview of our deep snapshot HDR imaging framework. It first estimates tentative HDR luminance and then estimates the HDR image in the luminance-normalized domain. The idea of luminance normalization enables to consider relative local contrasts in both bright and dark areas, regardless of the absolute luminance levels.}
\label{overview}
\end{figure*}

%===========================================================
\section{Proposed Deep Snapshot HDR Imaging}

%------------------------------------------------------------------------- 
\subsection{Framework overview}

In this paper, we assume the ME-CFA pattern shown in Fig.~\ref{fig:intro}, which consists of 4$\times$4 regular pattern with three exposure levels, assuming the mosaic of RGB filters combined with neutral density filters with three attenuation levels. Although our framework is general and not limited to this pattern, we use this because (i) it is based on the common Bayer pattern~\cite{bayer1976color}, similar to existing ME-CFA patterns~\cite{narasimhan2005enhancing,hajisharif2015adaptive}, and (ii) it consists of three exposures, which are commonly used in recent HDR imaging studies~\cite{kalantari2017deep,wu2018deep}. Those two conditions make it possible to experimentally compare our framework with standard Bayer demosaicking methods~\cite{cui2018color,kokkinos2018deep} and also state-of-the-art HDR imaging methods using three LDR images~\cite{kalantari2017deep,wu2018deep}.

Figure~\ref{overview} shows the overview of our framework, which mainly consists of two parts: (i) luminance estimation and (ii) luminance-normalized HDR image estimation. The first part estimates tentative HDR luminance based on the interpolated LDR images by the learned LDR-interpolation-network~(LDR-I-Net). Then, based on the tentative luminance, the second part estimates the luminance-normalized HDR image by the learned LN-Net. Each part is detailed in subsection~\ref{ssec:le} and~\ref{ssec:lne}, respectively. Finally, the HDR image is reconstructed by multiplying the tentative luminance and the estimated luminance-normalized HDR image.

Throughout this paper, we use the term ``irradiance'' or ``sensor irradiance~\cite{paul1997debevec}'' to represent the irradiance of the light reaching the image sensor after going through the camera's optical elements such as a lens. Because we assume a linear optical system as in~\cite{paul1997debevec,grossberg2003space}, the sensor irradiance is assumed to be the same as scene radiance in this paper. We also assume linear responses between the sensor irradiance and pixel values because we process the RAW data, which typically have linear camera responses.

%------------------------------------------------------------------------- 
\subsection{Luminance estimation}
\label{ssec:le}

In the luminance estimation, we first interpolate the missing RGB pixel values to generate interpolated LDR images. For this purpose, we train LDR-I-Net, for which we adopt the U-Net architecture~\cite{ronneberger2015u}. The inputs of LDR-I-Net are the ME-CFA RAW data and the linearly interpolated each color-exposure sample of the ME-CFA raw data. The loss function for LDR-I-Net is described as
\begin{eqnarray}
 \mathcal{L}_{\rm LDR} &=& \sum_{k=1}^{N} 
 ||{\bm f}_k\left({\bm y}; {\bm \theta}\right) - {\bm z}_k||_2^2
 \nonumber \\
 & &+ \lambda_{\rm LDR} 
 ||\nabla {\bm f}_k\left({\bm y}; {\bm \theta}\right) - \nabla {\bm z}_k||_2^2
 \,,
 \label{eq:LossLDRnet}
\end{eqnarray}
where
${\bm y} = [{\bm x}; {\bm h}({\bm x})]$ is the network input,
${\bm x}$ is the sub-mosaicked representation of the ME-CFA RAW data, as used in~\cite{henz2018deep} (e.g., sparse 16-channel data for the 4$\times$4 regular pattern), 
${\bm h}()$ represents the linear interpolation for the sparse data, 
${\bm f}_k({\bm y};{\bm \theta})$ is the output of LDR-I-Net for \mbox{$k$-th} exposure LDR image,
${\bm \theta}$ represents the network weights,
${\bm z}_k$ is true \mbox{$k$-th} exposure LDR image,
$N$ is the number of exposure levels in the ME-CFA,
$\nabla$ represents the horizontal and the vertical derivative operators,
and
$\lambda_{\rm LDR}$ is a hyper-parameter.
The second term, which we call the gradient term, evaluates the errors in the image gradient domain.
In this paper, we empirically set to $\lambda_{\rm LDR}=1$ for the hyper-parameter.
We will show the effectiveness of the gradient term by the ablation study in subsection~\ref{ss:Validation}.
Note that the loss function of Eq.~(\ref{eq:LossLDRnet}) evaluates the MSE in the LDR image domain, not in the HDR image domain. Thus, we use the standard MSE for loss computation.

Once we have interpolated LDR images, we apply Debevec's method~\cite{paul1997debevec} to the LDR images for tentative HDR image estimation. Then, tentative HDR luminance is derived as the maximum value of the RGB sensor irradiance values, which corresponds to the value~(V) in the HSV color space.
The tentative luminance can be formulated as
\begin{eqnarray}
 \hat{L}_i &=& \max_{c} \tilde{E}^{(c)}_i \,,
 \\
 \tilde{{\bm E}} &=& {\bm R}({\bm f}({\bm y}; {\bm \theta})) \,,
\end{eqnarray}
where
${\bm f}({\bm y}; {\bm \theta})$ is the interpolated LDR images,
${\bm R}()$ represents the Debevec's HDR image estimation operation~\cite{paul1997debevec},
$\tilde{{\bm E}}$ is the tentative HDR image estimated from the interpolated LDR images,
$\tilde{E}^{(c)}_i$ is the estimated tentative sensor irradiance of $c$-th channel at $i$-th pixel in $\tilde{{\bm E}}$, 
and $\hat{L}_i$ is the tentative luminance of $i$-th pixel, where $\max$ operation is performed in a pixel-by-pixel manner.

%------------------------------------------------------------------------- 
\subsection{Luminance-normalized HDR image estimation}
\label{ssec:lne}

In the luminance-normalized HDR image estimation, we train LN-Net to reconstruct the HDR image in the luminance-normalized domain. The inputs of LN-Net are the sub-mosaicked representation of the ME-CFA RAW data, its sensor irradiance version as we will explain below, and linearly interpolated versions of them. The irradiance data is normalized by the tentative luminance to consider relative local contrasts regardless of the absolute luminance levels. We detail each process to train LN-Net below.

\begin{figure}[t!]
    \centering
    \begin{minipage}[t]{0.25\hsize}
        \centering
        \includegraphics[width=\hsize,pagebox=cropbox,clip]{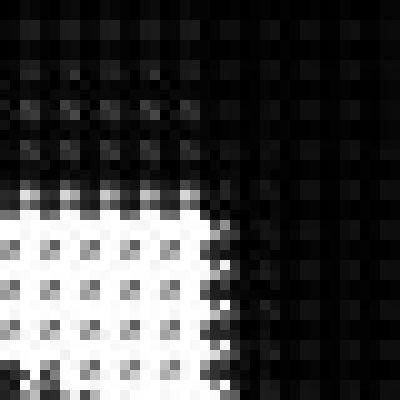}\\
        \small{(a)}
    \end{minipage}
    \begin{minipage}[t]{0.25\hsize}
        \centering
        \includegraphics[width=\hsize,pagebox=cropbox,clip]{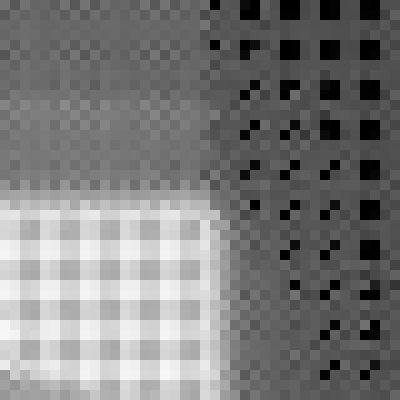}\\
        \small{(b)}
    \end{minipage}
    \begin{minipage}[t]{0.25\hsize}
        \centering
        \includegraphics[width=\hsize,pagebox=cropbox,clip]{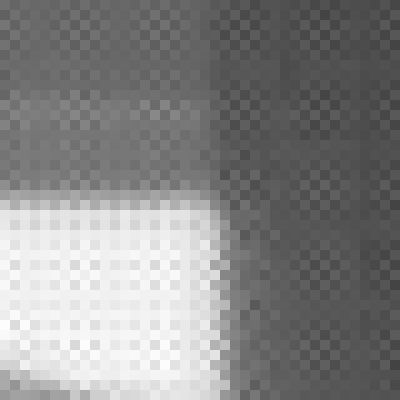}\\
        \small{(c)}
    \end{minipage}
    \caption{Examples of (a) the ME-CFA RAW data in the LDR domain, (b) the irradiance CFA RAW data converted by~Eq.~(\ref{eq:convert}), and (c) the irradiance CFA RAW data after the O/U-pixel correction.} \label{fig:handling}
\end{figure}

We first convert the ME-CFA RAW data to the sensor irradiance domain as
\begin{eqnarray}
 \xi_{k,i} &=& \frac{x_i}{\rho_k \Delta t} \,,
\label{eq:convert}
\end{eqnarray}
where
$x_{i}$ is $i$-th pixel value of the ME-CFA RAW data, 
$\rho_k$ is the attenuation factor for $k$-th exposure,
$\Delta t$ is the exposure time,
and
$\xi_{k,i}$ is the converted sensor irradiance of $i$-th pixel corresponding to $k$-th exposure.
In the snapshot case using an ME-CFA, the attenuation factor $\rho_k$ varies for each pixel according to the ME-CFA pattern, while the exposure time is constant for all pixels. Thus, in what follows, we set to $\Delta t = 1$, without loss of generality. Also, we call the irradiance data converted by~Eq.~(\ref{eq:convert}) ``irradiance CFA-RAW data'', in which different exposure levels are already corrected by converting the ME-CFA RAW data to the sensor irradiance domain. Fig.~\ref{fig:handling}(a) and \ref{fig:handling}(b) show the examples of the ME-CFA RAW data and the converted irradiance CFA RAW data.

% /////////////////////////////////////////////////////////////////////////

In the original ME-CFA RAW data, many pixels are over-exposed (saturated) or under-exposed (black-out) depending on the exposure level of each pixel compared to the scene radiance. Such over- or under-exposed pixels~(what we denote as ``O/U pixels'') have no meaningful irradiance information even after the conversion by~Eq.~(\ref{eq:convert}). Thus, we present an effective O/U-pixel correction method, which replaces the irradiance of an O/U pixel with the linearly interpolated value using adjacent lower- or higher-exposure irradiance samples. For example, the irradiance of an over-exposed pixel is corrected as
\begin{eqnarray}
\hat{\xi}_{k,i} &=&
\left\{
 \begin{array}{cc}
    \xi_{k,i}  & ( \xi_{k,i} \le \tau_{O,k} )  \\
    h_i({\bm \xi}_{k-1})  & ( \xi_{k,i} > \tau_{O,k} )
 \end{array}
\right.
\,,
\label{eq:OU}
\end{eqnarray}
where
the suffix $k$ represents $k$-th exposure,
the suffix $i$ represents $i$-th pixel,
$\hat{\xi}_{k,i}$ is the irradiance after the over-exposed pixel correction,
$\tau_{O,k}$ is the over-exposure threshold,
${\bm \xi}_{k-1}$ is the one-step lower-exposure sparse irradiance samples in the irradiance CFA RAW data,
and
$h_i({\bm \xi}_{k-1})$ is $i$-th pixel value of the linearly interpolated irradiance of ${\bm \xi}_{k-1}$.
We empirically set $0.995/(\rho_k \Delta t)$ to the over-exposure threshold $\tau_{O,k}$, where the range of the irradiance CFA RAW data is [0, 1].
This over-exposure correction is applied from the lower exposure data to the higher exposure data.
The under-exposed pixel correction is performed in the same manner, where the under-exposure threshold is set to $\tau_{U,k} = 0.005/(\rho_k \Delta t)$.
Figure~\ref{fig:handling}(b) and~\ref{fig:handling}(c) show the examples of the irradiance CFA RAW data before and after the O/U-pixel correction, respectively.

We then apply a linear interpolation to the corrected irradiance CFA RAW data to prepare the network input.
The corrected irradiance CFA RAW data,  $\hat{\bm \xi}$, and its linearly interpolated version, ${\bm h}(\hat{\bm \xi})$, can be considered as the HDR domain data, in which local contrasts in dark areas are very low compared with those in bright areas. Thus, we normalize the HDR domain data by the estimated tentative HDR luminance. This luminance normalization converts the absolute local contrasts to the relative local contrasts. We also use the LDR domain data of the sub-mosaicked ME-CFA RAW data ${\bm x}$ and its linearly interpolated version ${\bm h}({\bm x})$. For these LDR domain data, we do not perform the luminance normalization because the range of the absolute local contrasts is limited. 
The input to LN-Net ${\bm \eta}$ is described as
\begin{eqnarray}
 {\bm \eta} &=& \left[{\bm x}, {\bm h}({\bm x}), 
 \hat{\bm \xi}/\hat{\bm L}, {\bm h}(\hat{\bm \xi})/\hat{\bm L} \right]
 \,, \label{eq:eta}
\end{eqnarray}
where $\hat{\bm L}$ is the tentative luminance, and the divide operation is performed in a pixel-by-pixel manner.

We then estimate the luminance-normalized HDR image by LN-Net, where we adopt the U-Net like network architecture~\cite{ronneberger2015u}.
The loss function of LN-Net is the MSE in the luminance-normalized HDR domain as follows.
\begin{eqnarray}
 \mathcal{L}_{\rm LN} &=& 
 || {\bm g}({\bm \eta}; {\bm \psi}) - {\bm E}_{\rm LN} ||_2^2
 \nonumber \\
 & &+ \lambda_{\rm LN}
 ||\nabla {\bm g}({\bm \eta}; {\bm \psi}) - \nabla{\bm E}_{\rm LN} ||_2^2
 \,,
 \label{eq:LossLNnet}
\end{eqnarray}
where 
${\bm g}({\bm \eta}; {\bm \psi})$ represents LN-Net which estimates the luminance-normalized HDR image,
${\bm \psi}$ is the weights for the network,
${\bm \eta}$ is the network input defined in Eq.~(\ref{eq:eta}), 
${\bm E}_{\rm LN}$ is the luminance-normalized true HDR image,
$\nabla$ represents the horizontal and vertical gradient operator,
and
$\lambda_{\rm LN}$ is a hyper-parameter.
In this paper, we empirically set to $\lambda_{\rm LN}=1$ for the hyper-parameter.
The luminance-normalized true HDR image is defined by
\begin{eqnarray}
{\bm E}_{LN} &=& {\bm E}/\hat{\bm L} \,,
\label{eq:E_LN}
\end{eqnarray}
where ${\bm E}$ is the true HDR image and the divide operation is perform in a pixel-by-pixel manner.
We finally reconstruct the final HDR image as
\begin{eqnarray}
 \hat{\bm E} &=& \hat{\bm L} \times {\bm g}({\bm \eta}; {\bm \psi}) \,,
\label{eq:hat_E}
\end{eqnarray}
where the multiplying operation is performed in a pixel-by-pixel manner.

By substituting Eqs.~(\ref{eq:E_LN}) and~(\ref{eq:hat_E}) into Eq.~(\ref{eq:LossLNnet}), one can find that the loss function for LN-Net corresponds to the MSE normalized by the tentative luminance.
In the proposed luminance-normalized HDR image estimation, the input and the output of the network are normalized by the luminance, which enables us to consider the relative local contrasts, rather than the absolute differences.

%------------------------------------------------------------------------- 
\subsection{Network architecture}

In our framework, we use two networks: LDR-I-Net and LN-Net. In this paper, we adopt the U-Net architecture~\cite{ronneberger2015u} for both networks, where the depth of the U-Net is five.
Though, one can use any network architectures.

As mentioned above, the network inputs are a pair of the sparse sub-mosaicked data and the dense interpolated data. To adapt the data sparseness difference, we insert RAW data adaptation blocks as follows. The RAW data adaptation for the sparse data consists of a convolution layer with the ReLU activation whose kernel size is 7$\times$7.
The adaptation for the interpolated data is a convolution layer with the ReLU activation whose kernel size is $3\times3$. The outputs of both adaptations are concatenated and then fed into the network. We validate the effectiveness of the RAW data adaptation blocks in subsection~\ref{ss:Validation}. The detailed network architecture of the U-Net and the RAW data adaptation blocks are included in the supplemental material.

%===========================================================
\section{Experimental Results}

%------------------------------------------------------------------------- 
\subsection{Setups}

\noindent {\bf Datasets.} We used two public HDR image datasets for evaluating our framework: Funt's dataset~\cite{funt2010rehabilitation} and Kalantari's dataset~\cite{kalantari2017deep}.

Funt's dataset consists of static scenes. We used this dataset for comparison with other snapshot methods. Although other static HDR datasets are available as listed in~\cite{moriwaki2018hybrid}, we used Funt's dataset because it contains relatively a large number of HDR images~(224 images) generated using the same camera. Each HDR image is generated by Debevec's method~\cite{paul1997debevec} using 9~LDR images with the exposure value~(EV) set of \{-4, -3, -2, -1, 0, 1, 2, 3, 4\}.

Kalantari's dataset is recently used for evaluating dynamic scenes mainly with human motions. We used this dataset for comparison with state-of-the-art HDR imaging methods using a single or multiple LDR images. The dataset contains 89 dynamic scenes. For each scene, the ground-truth HDR image is generated using static LDR images taken with a reference human pose. In contrast, test LDR images are taken with a human motion including the reference pose. The EV set of \{-2, 0, 2\} or \{-3, 0, 3\} is used for the LDR image acquisition, where EV$=$0 is used to take the test LDR image with the reference pose.

\subsubsection{ME-CFA data simulation.} We simulated the ME-CFA data from the above-mentioned HDR data, which is 32-bit RGBE image format. We first normalized each HDR data to [0, 1] and scaled the normalized HDR data as [0, 1], [0, 4], and [0, 16], according to the three exposure levels corresponding to the assumed EV set of \{-2, 0, 2\}. Then, we clipped each scaled data by [0, 1] and quantized the clipped data by 8-bit depth. Finally, we sampled the quantized data according to the ME-CFA pattern to generate the mosaic ME-CFA data. By this data generation process, quantization errors were properly taken into account.

\subsubsection{Training setups.} For Funt's dataset, we used randomly selected 13 images for testing and the rest 211 images for training. For Kalantari's dataset, we used the provided 15 test and 74 training images. In the training phase, we randomly sampled 32$\times$32-sized patches from each training image set and randomly applied each of a horizontal flip, a vertical flip, and a swapping of horizontal and vertical axes (transpose) for data augmentation. The used optimizer is Adam~\cite{kingma2014adam}, where the learning rate was set to 0.001 and the parameters $(\beta_1, \beta_2)$ were set to $(0.9, 0.999)$. We performed 3,000 times mini-batch updates, where the mini-batch size was set to 32.

\subsubsection{Evaluation metrics.} We used the following five metrics: color PSNR~(CPSNR) in the linear HDR domain, CPSNR in the global tone-mapped domain~(G-CPSNR), CPSNR in the local tome-mapped domain~(L-CPSNR), HDR-VDP-2~\cite{mantiuk2011hdr}, and luminance-normalized MSE~(LN-MSE), which is the MSE normalized by the true luminance. We used the same global tone-mapping function as~\cite{kalantari2017deep} for G-CPSNR and the MATLAB local tone-mapping function for L-CPSNR. For each dataset, the average metric value of all test images is presented for comparison. For subjective evaluation, we used a commercial software, Photomatix\footnote{https://www.hdrsoft.com}, to apply local tone-mapping for effective visualization.

% /////////////////////////////////////////////////////////////////////////
\begin{table*}[t!]
    \centering
    \caption{Ablation study.} \label{ablationtable}
    \scalebox{0.9}{
    \begin{tabular}{|l||r|r|r|r|r|r|} \hline
         & CPSNR & G-CPSNR & L-CPSNR & HDR-VDP-2 & LN-MSE \\ \hline
        Ours (full version) & \bf{48.57} & \bf{41.94} & \bf{40.38} & \bf{80.57}  & \bf{0.0585}\\
        \quad without luminance normalization & 46.70 & 30.24 & 29.29 &  77.97 & 0.1289\\
        \quad without O/U-pixel correction & 42.30 & 36.60 & 34.45 & 78.51 & 0.0913 \\
        \quad without gradient term in the loss & 41.76 & 36.14 & 35.76 & 65.41 & 0.4364 \\
        \quad without RAW data adaptation & 46.19 & 39.45 & 37.77 & 78.16 & 0.0688\\ \hline
        \end{tabular}}
\end{table*}

%------------------------------------------------------------------------- 
\subsection{Validation study of our framework} \label{ss:Validation}

We first evaluate the validity of our framework using Funt's dataset. 

\subsubsection{Ablation study.} Table~\ref{ablationtable} shows the result for the ablation study. We can observe that the case without the normalization by luminance (second row) presents much lower G-PSNR and L-CPSNR values compared with the cases with the normalization. This is because the cases without the normalization tend to neglect local contrasts in dark areas, which decreases the performance of both the subjective and the objective evaluation in the tone-mapped domains, where dark areas are significantly enhanced. We can also observe that the other proposed components certainly contribute to the performance improvements in all evaluated metrics. 

\subsubsection{Loss domain comparison.} Table~\ref{losstable} shows the comparison of loss computation domains. The loss in the standard linear HDR domain presents lower G-CPSNR and L-CPSNR values because it tends to disregard the errors in dark areas. The loss in the global tone-mapped domain~\cite{kalantari2017deep} improves the G-CPSNR and the L-CPSNR performance, respectively. The loss in our proposed luminance-normalized domain provides further better performance by considering the relative local contrasts.

The above two studies demonstrate the effectiveness of our framework with the luminance normalization that enables effective local contrast consideration.

% /////////////////////////////////////////////////////////////////////////
\begin{table*}[t!]
    \centering
    \caption{Loss domain comparison.} \label{losstable}
    \scalebox{0.89}{
    \begin{tabular}{|l||r|r|r|r|r|r|} \hline
        Loss domain & CPSNR & G-CPSNR & L-CPSNR & HDR-VDP-2 & LN-MSE \\ \hline
        Linear HDR domain & 47.63 & 38.42 & 36.25 & 78.96 & 0.0749\\
        Global tone-mapped HDR domain & 47.56 & 41.57 & 39.57 & 79.73 & 0.0609 \\
        Our luminance-normalized HDR domain & \bf{48.57} & \bf{41.94} & \bf{40.38}  &  \bf{80.57} & \bf{0.0585}\\\hline
    \end{tabular}}
\end{table*}

%------------------------------------------------------------------------- 
\subsection{Comparison with other methods}

\subsubsection{Compared methods.} To the best of our knowledge, there is no publicly available source code that is directly workable and comparable for the considered snapshot HDR imaging problem. Thus, we compared our framework with two combination frameworks as follows.

The first one is the demosaicking-based framework. It first converts the ME-CFA RAW data to the irradiance CFA RAW data and then applies an existing Bayer demosaicking method to the irradiance CFA RAW data. To generate the irradiance CFA RAW data, we applied the same processes as in subsection~\ref{ssec:lne}, including our proposed O/U-pixel correction since it was confirmed that our pixel correction significantly improves the numerical performance of existing methods. We used state-of-the-art interpolation-based~(ARI~\cite{monno2017adaptive}) and deep learning-based~(Kokkions~\cite{kokkinos2018deep} and CDMNet~\cite{cui2018color}) demosaicking methods for comparison. 

The second one is the LDR-interpolation-based framework. It first interpolates (up-samples) the sub-mosaicked ME-CFA RAW data by an existing super-resolution~(SR) method with the scaling factor of~4 and then performs HDR reconstruction from the interpolated LDR images. We used existing competitive SR methods~(ESRGAN~\cite{wang2018esrgan}, WDSR~\cite{fan2018wide}, and EDSR~\cite{lim2017enhanced}) and our LDR-I-Net for SR and Debevec's method~\cite{paul1997debevec} for HDR reconstruction.

\subsubsection{\bf Results.} 
\begin{figure}[t!]
    \centering
    \includegraphics[width=\hsize,pagebox=cropbox,clip]{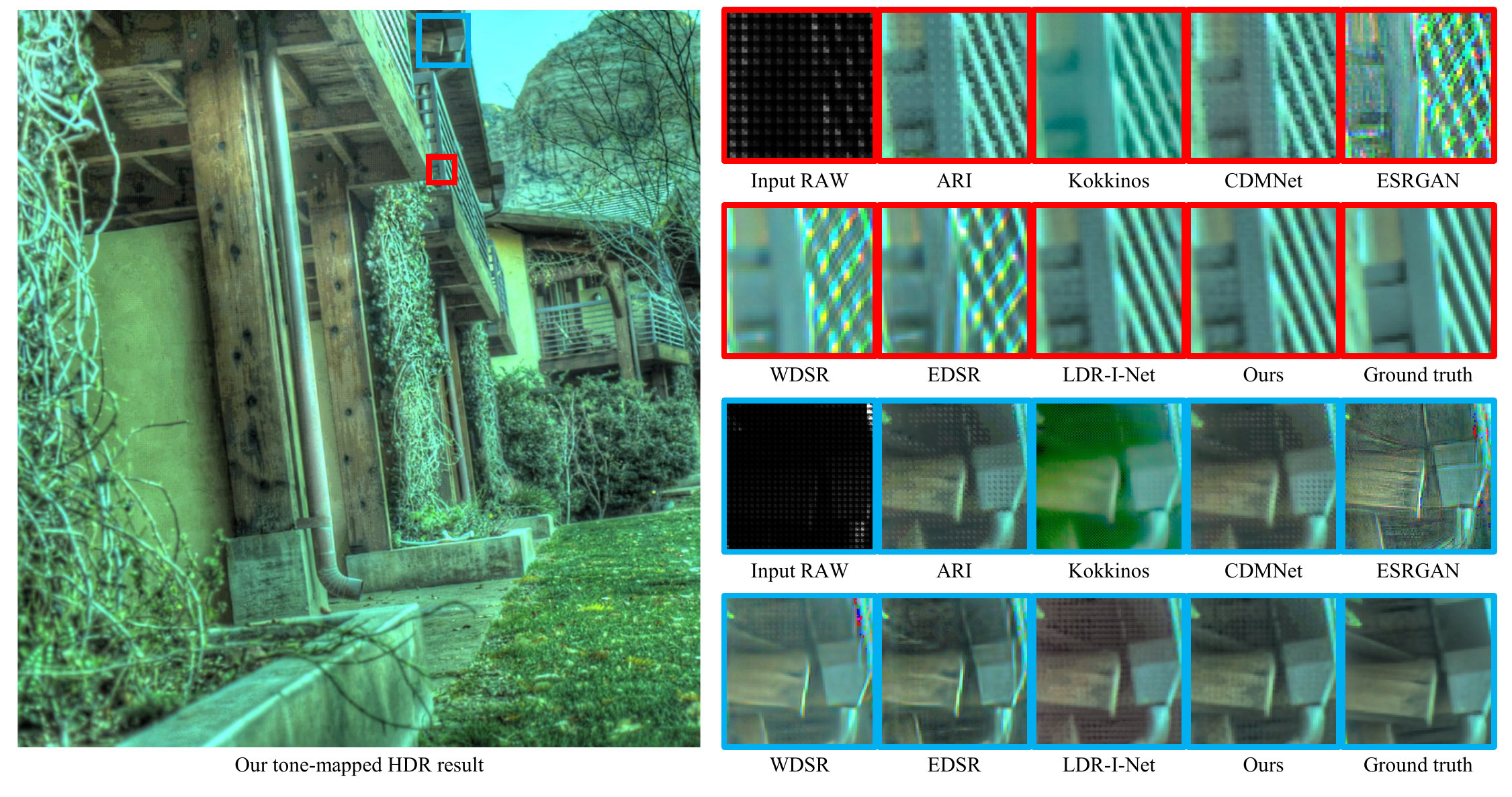}
    \caption{Visual comparisons on Funt's dataset.} \label{funtresultFig}
\end{figure}

Table~\ref{funtresult} and Fig.~\ref{funtresultFig} show the numerical and the visual comparisons using Funt's dataset. Table~\ref{funtresult} demonstrates that our framework can provide the best performance in all metrics. In Fig.~\ref{funtresultFig}, we can observe that the demosaicking-based methods generate severe zipper artifacts (ARI and CDMNet) or over-smoothed results (Kokkions), while the LDR-interpolation-based methods (ESRGAN, WDSR, and EDSR) generate severe aliasing artifacts for the high-frequency area of the red box. Although LDR-I-Net provides comparable results for bright areas, it generates severe quantization artifacts in the dark area of the blue box, because it only learns LDR interpolation, but does not perform any learning for HDR reconstruction. In contrast, our framework can produce a better result with fewer visible artifacts, though slight artifacts still appear in the dark area. More results can be seen in the supplemental material.

% /////////////////////////////////////////////////////////////////////////
\begin{table*}[t!]
    \begin{center}
    \caption{Comparison with two combination frameworks for snapshot HDR imaging.} \label{funtresult}
    \scalebox{0.8}{
    \begin{tabular}{|l|l||r|r|r|r|r|} \hline
        Framework & Demosaicking/SR & CPSNR & G-CPSNR & L-CPSNR & HDR-VDP-2 & LN-MSE \\ \hline
        \multirow{3}{*}{\begin{tabular}{l}\scriptsize{Demosaicking-based framework:}\\\scriptsize{Irradiance CFA RAW data generation}\\\scriptsize{$\rightarrow$ Demosaicking}\end{tabular}}
         & ARI~\cite{monno2017adaptive}  & 46.14 & 38.15 & 36.69 & 75.68 & 0.0712 \\
         & Kokkinos~\cite{kokkinos2018deep} & 41.06 & 26.27 & 26.65 & 69.32 & 0.1840 \\
         & CDMNet~\cite{cui2018color} & 46.32 & 38.37 & 37.12 & 58.00 & 0.0713 \\ \hline
        \multirow{4}{*}{\begin{tabular}{l}\scriptsize{LDR-interpolation-based framework:}\\\scriptsize{LDR interpolation by SR}\\\scriptsize{$\rightarrow$ HDR reconstruction}\end{tabular}}
         & ESRGAN~\cite{wang2018esrgan}  & 30.66 & 25.21 & 21.87 & 53.55 & 0.2720 \\
         & WDSR~\cite{fan2018wide} & 35.75 & 30.97 & 29.32 & 60.16 & 0.3796 \\
         & EDSR~\cite{lim2017enhanced} & 39.19 & 32.57 & 29.95 & 66.04 & 0.1190 \\ 
         & LDR-I-Net & 43.38 & 35.64 & 34.54 & 76.30 & 0.1030 \\\hline
        \multicolumn{2}{|l||}{Our deep snapshot HDR imaging framework} & \bf{48.57} & \bf{41.94}  & \bf{40.38} & \bf{80.57} & \bf{0.0585}\\\hline
    \end{tabular}}
  \end{center}
\end{table*}

% /////////////////////////////////////////////////////////////////////////

%------------------------------------------------------------------------- 
\subsection{Comparison with state-of-the-art methods using a single or multiple LDR images}

\subsubsection{Compared methods.} Using Kalantari's dataset, we next compare our snapshot framework with state-of-the-art HDR imaging methods using multiple LDR images (Sen~\cite{sen2012robust}, Kalantari~\cite{kalantari2017deep}, and Wu~\cite{wu2018deep}) or a single LDR image~(HDRCNN~\cite{eilertsen2017hdr}, DrTMO~\cite{endo2017deep}, and ExpandNet~\cite{marnerides2018expandnet}). We used all three LDR images for the multi-LDR-based methods and the second-exposure LDR image for the single-LDR-based methods.

% /////////////////////////////////////////////////////////////////////////
\begin{table*}[t!]
    \begin{center}
    \caption{Comparison with state-of-the-art HDR imaging methods.}\label{kalantsariresult}
    \scalebox{0.9}{    
    \begin{tabular}{|l|l||r|r|r|r|r|} \hline
        Input sources & Methods & CPSNR & G-CPSNR & L-CPSNR & HDR-VDP-2 & LN-MSE \\ \hline
        \multirow{3}{*}{Multiple LDR images}
         & Sen~\cite{sen2012robust} & 38.05 & 40.76 & 36.13 & 61.08 & 0.0389 \\
         & Kalantari~\cite{kalantari2017deep} &  41.15 & \bf{42.65} & \bf{38.22} & 64.57 & \bf{0.0306}\\
         & Wu~\cite{wu2018deep} & 40.88 & 42.53 & 37.98 & 65.60 & 0.0338 \\\hline
        \multirow{3}{*}{\begin{tabular}{l}Single LDR image\\(Second exposure)\end{tabular}}
         & HDRCNN~\cite{eilertsen2017hdr} & 12.92 & 14.13 & 34.80 & 54.48 & 4.1082\\
         & DrTMO~\cite{endo2017deep} & 18.23 & 14.07  & 25.32 & 56.78 & 8.7912\\
         & ExpandNet~\cite{marnerides2018expandnet} & 22.09 & 22.37 & 28.03 & 57.34 & 1.2923 \\ \hline
        ME-CFA RAW data & Ours & \bf{41.43} & 38.60  & 35.23 &  \bf{66.59} & 0.0832 \\ \hline
        \end{tabular}}
  \end{center}
\end{table*}

% /////////////////////////////////////////////////////////////////////////
\begin{figure*}[t!]
    \centering
    \includegraphics[width=12cm,pagebox=cropbox,clip]{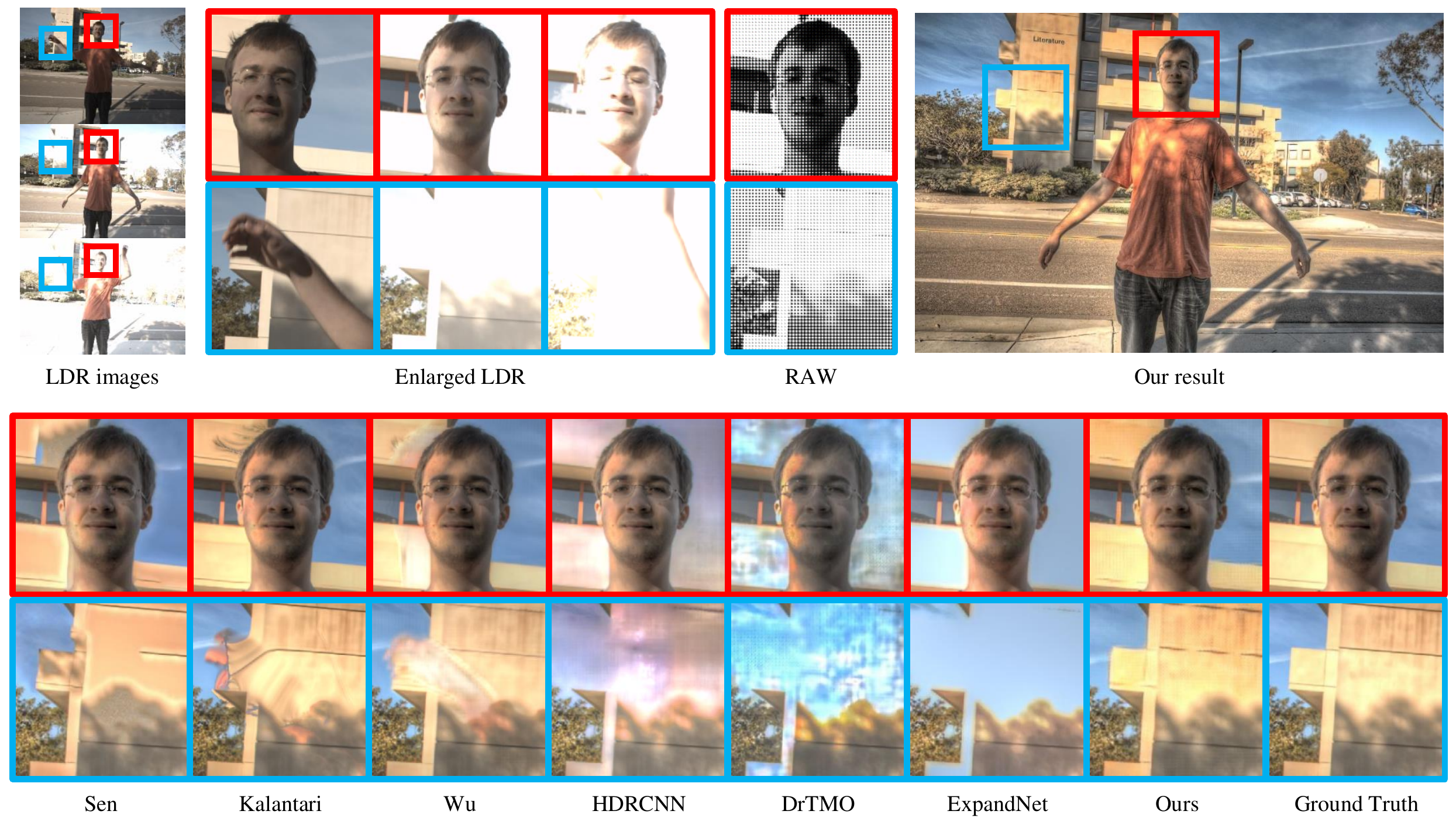}
    \caption{Visual comparisons on Kalantari's dataset.} \label{kalantsariresultFig}
\end{figure*}

\subsubsection{Results.} Figure~\ref{kalantsariresultFig} shows the visual comparison. We can observe that the multi-LDR-based methods (Sen, Kalantari, and Wu) generate severe ghost artifacts in the red and the blue box areas, which are due to the head and the arm motions between input LDR images. The single-LDR-based methods~(HDRCNN, DrTMO, and ExpandNet) generate severe inpainting artifacts for the over-exposed areas in the input second-exposure LDR image. In contrast, our framework can reconstruct a visually pleasing result with much fewer visible artifacts. More results can be seen in the supplemental material.

\begin{figure*}[t!]
    \centering
    \includegraphics[width=12cm,pagebox=cropbox,clip]{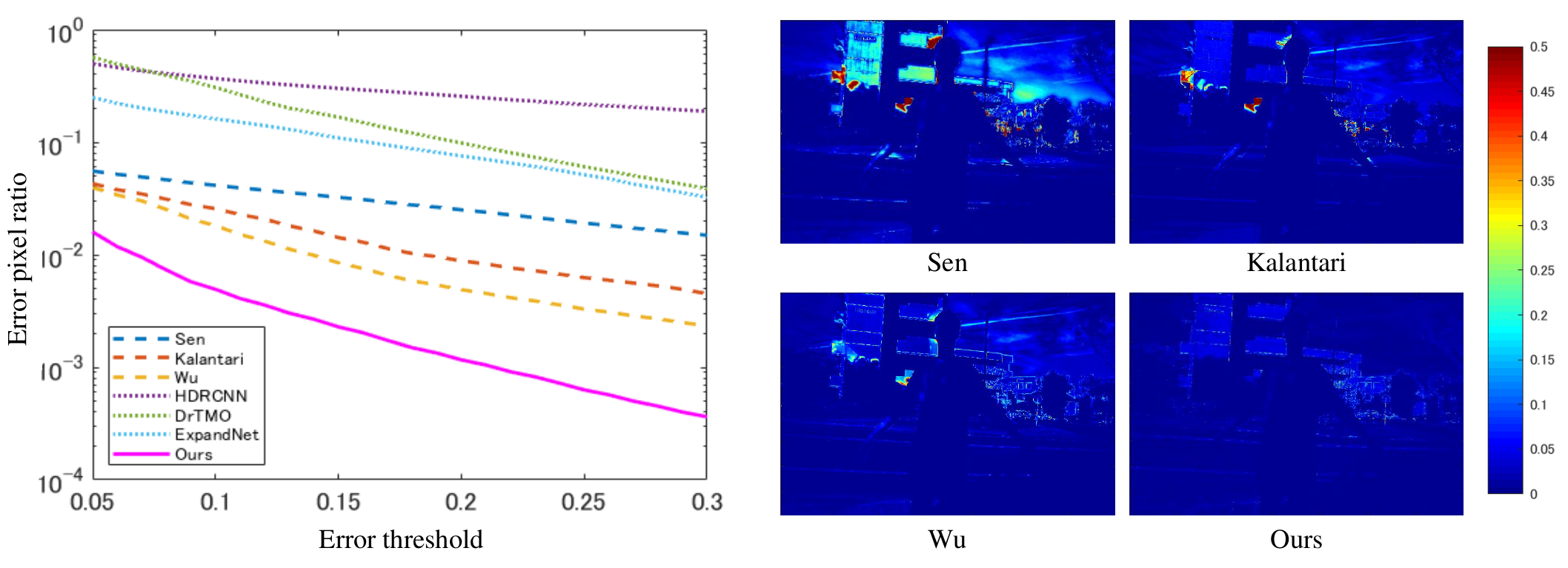}
    \caption{Left: Comparison of the error pixel ratio. Right: Comparison of the error maps.}
    \label{kalantsariresultep}
\end{figure*}

Table~\ref{kalantsariresult} shows the numerical comparison. Our framework provides the highest score in CPSNR and HDR-VDP-2. In contrast, the multi-LDR-based methods present better performance for the other metrics. This is because these methods have the benefit of having all three-exposure information for each pixel, and thus should provide better performance for static regions, which are dominant in each scene of Kalantari's dataset. However, as shown in the visual comparison, these methods are very susceptible to ghost artifacts, which significantly disturbs visual perception and makes the perceptual HDR-VDP-2 score lower. To quantitatively evaluate such significant artifacts, in Fig.~\ref{kalantsariresultep}, we evaluate the ratio of error pixels whose MSE of RGB irradiance values is larger than the threshold of the horizontal axis. From the result, we can clearly observe that our snapshot framework can generate the HDR image with much fewer error pixels. We also show the comparison of error maps, where the multi-LDR-based methods generate significant errors in the dynamic regions around the head and the arm.

% /////////////////////////////////////////////////////////////////////////

%------------------------------------------------------------------------- 
\subsection{Limitation}

In our results, zipper artifacts still remain in some areas. This is because of the very challenging nature of snapshot HDR reconstruction with very sparse sampling of each color-exposure component and many saturated/blacked-out pixels. Furthermore, in the snapshot HDR problem, zipper artifacts may occur even for uniform areas without textures because of the differences of the quantization levels of three exposure images, meaning that converted sensor irradiance values in the uniform area do not match completely among the three exposure levels. To reduce the remaining zipper artifacts is our future work.

%===========================================================
\section{Conclusion}

In this paper, we have proposed a novel deep learning-based framework that can effectively address the joint demosaicking and HDR reconstruction problem for snapshot HDR imaging using an ME-CFA. We have introduced the idea of luminance normalization that simultaneously enables effective loss computation and input data normalization to learn the HDR image reconstruction network by considering relative local image contrasts. Experimental results have demonstrated that our framework can produce HDR images with much fewer visual artifacts compared with other snapshot methods and also state-of-the-art HDR imaging methods using multiple LDR images. We provide the source code to reproduce our results at \textcolor{blue}{http://www.ok.sc.e.titech.ac.jp/res/DSHDR/index.html}.

%===========================================================
\bibliographystyle{splncs}
\bibliography{egbib}

\begin{thebibliography}{10}

\bibitem{paul1997debevec}
Debevec, P., Malik, J.:
\newblock Recovering high dynamic range radiance maps from photographs.
\newblock In: Proc. of SIGGRAPH. (1997)  1--10

\bibitem{kalantari2017deep}
Kalantari, N.K., Ramamoorthi, R.:
\newblock Deep high dynamic range imaging of dynamic scenes.
\newblock ACM Trans. on Graphics \textbf{36} (2017)  144:1--12

\bibitem{wu2018deep}
Wu, S., Xu, J., Tai, Y.W., Tang, C.K.:
\newblock Deep high dynamic range imaging with large foreground motions.
\newblock In: Proc. of European Conf. on Computer Vision (ECCV). (2018)
  121--135

\bibitem{yan2019attention}
Yan, Q., Gong, D., Shi, Q., van~den Hengel, A., Shen, C., Reid, I., Zhang, Y.:
\newblock Attention-guided network for ghost-free high dynamic range imaging.
\newblock In: Proc. of IEEE Conf. on Computer Vision and Pattern Recognition
  (CVPR). (2019)  1751--1760

\bibitem{marnerides2018expandnet}
Marnerides, D., Bashford-Rogers, T., Hatchett, J., Debattista, K.:
\newblock {ExpandNet}: A deep convolutional neural network for high dynamic
  range expansion from low dynamic range content.
\newblock Computer Graphics Forum \textbf{37} (2018)  37--49

\bibitem{eilertsen2017hdr}
Eilertsen, G., Kronander, J., Denes, G., Mantiuk, R.K., Unger, J.:
\newblock {HDR} image reconstruction from a single exposure using deep {CNNs}.
\newblock ACM Trans. on Graphics \textbf{36} (2017)  178:1--15

\bibitem{lee2018deepc}
Lee, S., An, G.H., Kang, S.J.:
\newblock Deep chain {HDRI}: Reconstructing a high dynamic range image from a
  single low dynamic range image.
\newblock IEEE Access \textbf{6} (2018)  49913--49924

\bibitem{cho2014single}
Cho, H., Kim, S.J., Lee, S.:
\newblock Single-shot high dynamic range imaging using coded electronic
  shutter.
\newblock Computer Graphics Forum \textbf{33} (2014)  329--338

\bibitem{choi2017reconstructing}
Choi, I., Baek, S.H., Kim, M.H.:
\newblock Reconstructing interlaced high-dynamic-range video using joint
  learning.
\newblock IEEE Trans. on Image Processing \textbf{26} (2017)  5353--5366

\bibitem{narasimhan2005enhancing}
Narasimhan, S.G., Nayar, S.K.:
\newblock Enhancing resolution along multiple imaging dimensions using assorted
  pixels.
\newblock IEEE Trans. on Pattern Analysis and Machine Intelligence \textbf{27}
  (2005)  518--530

\bibitem{nayar2000high}
Nayar, S.K., Mitsunaga, T.:
\newblock High dynamic range imaging: Spatially varying pixel exposures.
\newblock In: Proc. of IEEE Conf. on Computer Vision and Pattern Recognition
  (CVPR). (2000)  1--8

\bibitem{eilertsen2015real}
Eilertsen, G., Mantiuk, R.K., Unger, J.:
\newblock Real-time noise-aware tone mapping.
\newblock ACM Trans. on Graphics \textbf{34} (2015)  198:1--15

\bibitem{ma2017multi}
Ma, K., Duanmu, Z., Yeganeh, H., Wang, Z.:
\newblock Multi-exposure image fusion by optimizing a structural similarity
  index.
\newblock IEEE Trans. on Computational Imaging \textbf{4} (2017)  60--72

\bibitem{ma2017robust}
Ma, K., Li, H., Yong, H., Wang, Z., Meng, D., Zhang, L.:
\newblock Robust multi-exposure image fusion: a structural patch decomposition
  approach.
\newblock IEEE Trans. on Image Processing \textbf{26} (2017)  2519--2532

\bibitem{mertens2009exposure}
Mertens, T., Kautz, J., Van~Reeth, F.:
\newblock Exposure fusion: A simple and practical alternative to high dynamic
  range photography.
\newblock Computer Graphics Forum \textbf{28} (2009)  161--171

\bibitem{hu2013hdr}
Hu, J., Gallo, O., Pulli, K., Sun, X.:
\newblock {HDR} deghosting: How to deal with saturation?
\newblock In: Proc. of IEEE Conf. on Computer Vision and Pattern Recognition
  (CVPR). (2013)  1163--1170

\bibitem{sen2012robust}
Sen, P., Kalantari, N.K., Yaesoubi, M., Darabi, S., Goldman, D.B., Shechtman,
  E.:
\newblock Robust patch-based {HDR} reconstruction of dynamic scenes.
\newblock ACM Trans. on Graphics \textbf{31} (2012)  203:1--11

\bibitem{lee2014ghost}
Lee, C., Li, Y., Monga, V.:
\newblock Ghost-free high dynamic range imaging via rank minimization.
\newblock IEEE Signal Processing Letters \textbf{21} (2014)  1045--1049

\bibitem{oh2014robust}
Oh, T.H., Lee, J.Y., Tai, Y.W., Kweon, I.S.:
\newblock Robust high dynamic range imaging by rank minimization.
\newblock IEEE Trans. on Pattern Analysis and Machine Intelligence \textbf{37}
  (2014)  1219--1232

\bibitem{hasinoff2016burst}
Hasinoff, S.W., Sharlet, D., Geiss, R., Adams, A., Barron, J.T., Kainz, F.,
  Chen, J., Levoy, M.:
\newblock Burst photography for high dynamic range and low-light imaging on
  mobile cameras.
\newblock ACM Trans. on Graphics \textbf{35} (2016)  192:1--12

\bibitem{kalantari2019deep}
Kalantari, N.K., Ramamoorthi, R.:
\newblock Deep {HDR} video from sequences with alternating exposures.
\newblock Computer Graphics Forum \textbf{38} (2019)  193--205

\bibitem{prabhakar2019fast}
Prabhakar, K.R., Arora, R., Swaminathan, A., Singh, K.P., Babu, R.V.:
\newblock A fast, scalable, and reliable deghosting method for extreme exposure
  fusion.
\newblock In: Proc. of IEEE Int. Conf. on Computational Photography (ICCP).
  (2019)  170--177

\bibitem{ram2017deepfuse}
Ram~Prabhakar, K., Sai~Srikar, V., Venkatesh~Babu, R.:
\newblock Deep{F}use: A deep unsupervised approach for exposure fusion with
  extreme exposure image pairs.
\newblock In: Proc. of IEEE Int. Conf. on Computer Vision (ICCV). (2017)
  4724--4732

\bibitem{yan2019multi}
Yan, Q., Gong, D., Zhang, P., Shi, Q., Sun, J., Reid, I., Zhang, Y.:
\newblock Multi-scale dense networks for deep high dynamic range imaging.
\newblock In: Proc. of IEEE Winter Conf. on Applications of Computer Vision
  (WACV). (2019)  41--50

\bibitem{tursun2015state}
Tursun, O.T., Aky{\"u}z, A.O., Erdem, A., Erdem, E.:
\newblock The state of the art in {HDR} deghosting: A survey and evaluation.
\newblock Computer Graphics Forum \textbf{34} (2015)  683--707

\bibitem{ogino2016super}
Ogino, Y., Tanaka, M., Shibata, T., Okutomi, M.:
\newblock Super high dynamic range video.
\newblock In: Proc. of Int. Conf. on Pattern Recognition (ICPR). (2016)
  4208--4213

\bibitem{tocci2011versatile}
Tocci, M.D., Kiser, C., Tocci, N., Sen, P.:
\newblock A versatile {HDR} video production system.
\newblock ACM Trans. on Graphics \textbf{30} (2011)  41:1--9

\bibitem{yang2018image}
Yang, X., Xu, K., Song, Y., Zhang, Q., Wei, X., Lau, R.W.:
\newblock Image correction via deep reciprocating {HDR} transformation.
\newblock In: Proc. of IEEE Conf. on Computer Vision and Pattern Recognition
  (CVPR). (2018)  1798--1807

\bibitem{moriwaki2018hybrid}
Moriwaki, K., Yoshihashi, R., Kawakami, R., You, S., Naemura, T.:
\newblock Hybrid loss for learning single-image-based {HDR} reconstruction.
\newblock arXiv preprint 1812.07134 (2018)

\bibitem{kim2018itm}
Kim, S.Y., Kim, D.E., Kim, M.:
\newblock {ITM-CNN}: Learning the inverse tone mapping from low dynamic range
  video to high dynamic range displays using convolutional neural networks.
\newblock In: Proc. of Asian Conf. on Computer Vision (ACCV). (2018)  395--409

\bibitem{endo2017deep}
Endo, Y., Kanamori, Y., Mitani, J.:
\newblock Deep reverse tone mapping.
\newblock ACM Trans. on Graphics \textbf{36} (2017)  177:1--10

\bibitem{lee2018deepr}
Lee, S., Hwan~An, G., Kang, S.J.:
\newblock Deep recursive {HDRI}: Inverse tone mapping using generative
  adversarial networks.
\newblock In: Proc. of European Conf. on Computer Vision (ECCV). (2018)
  613--628

\bibitem{gu2010coded}
Gu, J., Hitomi, Y., Mitsunaga, T., Nayar, S.:
\newblock Coded rolling shutter photography: Flexible space-time sampling.
\newblock In: Proc. of IEEE Int. Conf. on Computational Photography (ICCP).
  (2010)  1--8

\bibitem{uda2016variable}
Uda, S., Sakaue, F., Sato, J.:
\newblock Variable exposure time imaging for obtaining unblurred {HDR} images.
\newblock IPSJ Trans. on Computer Vision and Applications \textbf{8} (2016)
  3:1--7

\bibitem{alghamdi2019reconfigurable}
Alghamdi, M., Fu, Q., Thabet, A., Heidrich, W.:
\newblock Reconfigurable snapshot {HDR} imaging using coded masks and inception
  network.
\newblock In: Proc. of Vision, Modeling, and Visualization~(VMV). (2019)  1--9

\bibitem{nagahara2018space}
Nagahara, H., Sonoda, T., Liu, D., Gu, J.:
\newblock Space-time-brightness sampling using an adaptive pixel-wise coded
  exposure.
\newblock In: Proc. of IEEE Conf. on Computer Vision and Pattern Recognition
  Workshops (CVPRW). (2018)  1834--1842

\bibitem{serrano2016convolutional}
Serrano, A., Heide, F., Gutierrez, D., Wetzstein, G., Masia, B.:
\newblock Convolutional sparse coding for high dynamic range imaging.
\newblock Computer Graphics Forum \textbf{35} (2016)  153--163

\bibitem{go2018image}
Go, C., Kinoshita, Y., Shiota, S., Kiua, H.:
\newblock Image fusion for single-shot high dynamic range imaging with
  spatially varying exposures.
\newblock In: Proc. of Asia-Pacific Signal and Information Processing
  Association Annual Summit and Conference (APSIPA ASC). (2018)  1082--1086

\bibitem{hajisharif2015adaptive}
Hajisharif, S., Kronander, J., Unger, J.:
\newblock Adaptive dual{ISO} {HDR} reconstruction.
\newblock EURASIP Journal on Image and Video Processing \textbf{2015} (2015)
  1--13

\bibitem{heide2014flexisp}
Heide, F., Steinberger, M., Tsai, Y.T., Rouf, M., Paj{\k{a}}k, D., Reddy, D.,
  Gallo, O., Liu, J., Heidrich, W., Egiazarian, K., Kautz, J., Pulli, K.:
\newblock Flex{ISP}: A flexible camera image processing framework.
\newblock ACM Trans. on Graphics \textbf{33} (2014)  231:1--13

\bibitem{aguerrebere2017bayesian}
Aguerrebere, C., Almansa, A., Delon, J., Gousseau, Y., Mus{\'e}, P.:
\newblock A {B}ayesian hyperprior approach for joint image denoising and
  interpolation, with an application to {HDR} imaging.
\newblock IEEE Trans. on Computational Imaging \textbf{3} (2017)  633--646

\bibitem{aguerrebere2014single}
Aguerrebere, C., Almansa, A., Gousseau, Y., Delon, J., Muse, P.:
\newblock Single shot high dynamic range imaging using piecewise linear
  estimators.
\newblock In: Proc. of IEEE Int. Conf. on Computational Photography (ICCP).
  (2014)  1--10

\bibitem{an2017single}
An, V.G., Lee, C.:
\newblock Single-shot high dynamic range imaging via deep convolutional neural
  network.
\newblock In: Proc. of Asia-Pacific Signal and Information Processing
  Association Annual Summit and Conference (APSIPA ASC). (2017)  1768--1772

\bibitem{rouf2018high}
Rouf, M., Ward, R.K.:
\newblock High dynamic range imaging with a single exposure-multiplexed image
  using smooth contour prior.
\newblock In: Proc. of IS\&T Int. Symposium on Electronic Imaging (EI). (2018)
  440:1--6

\bibitem{cheng2009high}
Cheng, C.H., Au, O.C., Cheung, N.M., Liu, C.H., Yip, K.Y.:
\newblock High dynamic range image capturing by spatial varying exposed color
  filter array with specific demosaicking algorithm.
\newblock In: Proc. of IEEE Pacific Rim Conf. on Communications, Computers and
  Signal Processing (PACRIM). (2009)  648--653

\bibitem{bayer1976color}
Bayer, B.E.:
\newblock Color imaging array, {US} patent 3,971,065 (1976)

\bibitem{cui2018color}
Cui, K., Jin, Z., Steinbach, E.:
\newblock Color image demosaicking using a 3-stage convolutional neural network
  structure.
\newblock In: Proc. of IEEE Int. Conf. on Image Processing (ICIP). (2018)
  2177--2181

\bibitem{kokkinos2018deep}
Kokkinos, F., Lefkimmiatis, S.:
\newblock Deep image demosaicking using a cascade of convolutional residual
  denoising networks.
\newblock In: Proc. of European Conf. on Computer Vision (ECCV). (2018)
  317--333

\bibitem{grossberg2003space}
Grossberg, M.D., Nayar, S.K.:
\newblock What is the space of camera response functions?
\newblock In: Proc. of IEEE Conf. on Computer Vision and Pattern Recognition
  (CVPR). (2003)  1--8

\bibitem{ronneberger2015u}
Ronneberger, O., Fischer, P., Brox, T.:
\newblock U-net: Convolutional networks for biomedical image segmentation.
\newblock In: Proc. of Int. Conf. on Medical Image Computing and
  Computer-Assisted Intervention (MICCAI). (2015)  234--241

\bibitem{henz2018deep}
Henz, B., Gastal, E.S., Oliveira, M.M.:
\newblock Deep joint design of color filter arrays and demosaicing.
\newblock Computer Graphics Forum \textbf{37} (2018)  389--399

\bibitem{funt2010rehabilitation}
Funt, B., Shi, L.:
\newblock The rehabilitation of {MaxRGB}.
\newblock In: Proc. of Color and Imaging Conference (CIC). (2010)  256--259

\bibitem{kingma2014adam}
Kingma, D.P., Ba, J.:
\newblock Adam: A method for stochastic optimization.
\newblock arXiv preprint 1412.6980 (2014)

\bibitem{mantiuk2011hdr}
Mantiuk, R., Kim, K.J., Rempel, A.G., Heidrich, W.:
\newblock {HDR-VDP-2}: A calibrated visual metric for visibility and quality
  predictions in all luminance conditions.
\newblock ACM Trans. on Graphics \textbf{30} (2011)  40:1--13

\bibitem{monno2017adaptive}
Monno, Y., Kiku, D., Tanaka, M., Okutomi, M.:
\newblock Adaptive residual interpolation for color and multispectral image
  demosaicking.
\newblock Sensors \textbf{17} (2017)  2787:1--21

\bibitem{wang2018esrgan}
Wang, X., Yu, K., Wu, S., Gu, J., Liu, Y., Dong, C., Qiao, Y., Change~Loy, C.:
\newblock {ESRGAN}: Enhanced super-resolution generative adversarial networks.
\newblock In: Proc. of European Conf. on Computer Vision Workshops (ECCVW).
  (2018)  1--16

\bibitem{fan2018wide}
Fan, Y., Yu, J., Huang, T.S.:
\newblock Wide-activated deep residual networks based restoration for
  {BPG}-compressed images.
\newblock In: Proc. of IEEE Conf. on Computer Vision and Pattern Recognition
  workshops (CVPRW). (2018)  2621--2624

\bibitem{lim2017enhanced}
Lim, B., Son, S., Kim, H., Nah, S., Mu~Lee, K.:
\newblock Enhanced deep residual networks for single image super-resolution.
\newblock In: Proc. of IEEE Conf. on Computer Vision and Pattern Recognition
  Workshops (CVPRW). (2017)  1132--1140

\end{thebibliography}

\end{document}